%% file: tmm.tex
\documentclass[lettersize,journal]{IEEEtran}
\usepackage{amsmath,amsfonts}
\usepackage{algorithm}
\usepackage{array}
\usepackage[caption=false,font=normalsize,labelfont=sf,textfont=sf]{subfig}
\usepackage{textcomp}
\usepackage{stfloats}
\usepackage{url}
\usepackage{verbatim}
\usepackage{graphicx}
\usepackage{cite}
\usepackage[hidelinks]{hyperref}
\usepackage{multicol}
\usepackage{multirow}
\usepackage[dvipsnames]{xcolor}
\definecolor{high}{HTML}{26408B}
\definecolor{mid}{HTML}{81B1D5}
\usepackage{balance}
\usepackage{amsthm}
\usepackage{amssymb}
\usepackage[switch]{lineno}
\input{math-extension.tex}
\begin{document}

\title{GaussianForest: Hierarchical-Hybrid 3D Gaussian Splatting for Compressed Scene Modeling}

\author{Fengyi Zhang, Yadan Luo, Tianjun Zhang, Lin Zhang~\IEEEmembership{Senior Member,~IEEE},
Zi Huang~\IEEEmembership{Senior Member,~IEEE} 

\thanks{
Fengyi Zhang, Tianjun Zhang, and Lin Zhang are with the School of Software Engineering, Tongji University, Shanghai 201804, China (email: \{zzfff, 1911036, cslinzhang\}@tongji.edu.cn).

Yadan Luo and Zi Huang are with the University of Queensland, Brisbane, QLD 4072, Australia (e-mail: \{y.luo, helen.huang\}@uq.edu.au).
}
}



\maketitle

\begin{abstract}
The field of novel-view synthesis has recently witnessed the emergence of 3D Gaussian Splatting, which represents scenes in a point-based manner and renders through rasterization. This methodology, in contrast to Radiance Fields that rely on ray tracing, demonstrates superior rendering quality and speed.  However, the explicit and unstructured nature of 3D Gaussians poses a significant storage challenge, impeding its broader application. To address this challenge, we introduce the GaussianForest modeling framework, which hierarchically represents a scene as a forest of hybrid 3D Gaussians. Each hybrid Gaussian retains its unique explicit attributes while sharing implicit ones with its sibling Gaussians, thus optimizing parameterization with significantly fewer variables. Moreover, adaptive tree growth and pruning strategies are designed, ensuring detailed representation in complex regions and a notable reduction in the number of required Gaussians. Extensive experiments demonstrate that GaussianForest not only maintains comparable speed and quality but also achieves a compression rate surpassing 10 times, marking a significant advancement in efficient scene modeling. 
Codes will be available at \url{https://github.com/Xian-Bei/GaussianForest}.
\end{abstract}


\section{Introduction}

\IEEEPARstart{O}{ver} the past few years, there has been rapid development in the field of 3D vision, marked by the emergence of the Radiance Field technique designed for 3D scene representation and novel view synthesis. This development has not only established a solid foundation but also acted as a significant catalyst for further advancements.
As a pioneering effort, NeRF \cite{nerf} represents 3D scenes implicitly using Multi-Layer Perceptrons (MLPs) and employs ray tracing for rendering, resulting in high visual quality. 
However, this approach comes with the drawback of unacceptably slow speeds for both training and inference.
Subsequent research endeavors have explored various explicit or hybrid scene representations to enhance computational efficiency. Nonetheless, as these methods continue relying on ray tracing, which necessitates dense sampling across thousands of rays even in empty spaces, they encounter challenges in achieving real-time rendering rates. This challenge becomes more prominent when facing practical requirements such as high resolution, large-scale scenes, and consumer-grade devices.

\begin{figure}[ht]
    \centering
    \includegraphics[width=0.9\linewidth]{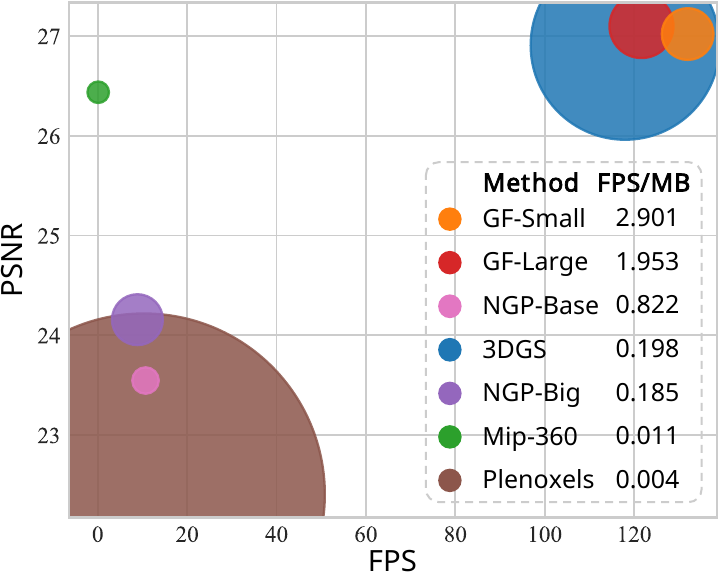}
    \caption{
    Quantitative comparison across 13 real-world scenes from three datasets 
    on rendering quality, model size, and rendering speed. 
    The size of each point in the figure indicates the corresponding model size (in MB). 
    Our GaussianForest (GF) excels in adeptly balancing rendering speed and model size. Across all scenarios, GF achieves the highest speed-to-size ratio, surpassing all baselines by a large margin while ensuring high-fidelity rendering quality. 
}
    \label{fig:sota}
\end{figure}

As a recent revolutionary development, 3D Gaussian Splatting (3DGS) \cite{3dgs} has introduced an explicitly point-based approach for scene representation, pivoting from ray tracing to rasterization for both training and rendering processes. This innovative shift has resulted in state-of-the-art visual quality and comparable training efficiency while significantly boosting rendering speed.
However, it comes with a primary constraint, which lies in its substantial storage requirements. 
Typically, it necessitates millions of Gaussians to represent a scene, resulting in a huge model size that even reaches thousands of megabytes.
Such resource-intensive demands pose a significant obstacle to its practical application, particularly in scenarios with limited resources and bandwidth.

In response to this practical challenge, we propose GaussianForest for compressed 3D scene representation, which models each Gaussian with significantly fewer parameters by organizing hybrid Gaussians in a hierarchical forest structure, while concurrently controlling their overall number via adaptive growth and pruning.
Our approach is motivated by the substantial parameter redundancy observed among the millions of Gaussians employed in 3DGS \cite{3dgs}, where groups of Gaussians exhibit implicit associations and share similar attributes. 
As demonstrated in Fig. \ref{fig:sota}, GaussianForest adeptly balances storage, speed, and rendering quality. 
The success of GaussianForest hinges on three pivotal elements.

Firstly, we introduce a hybrid representation of 3D Gaussians, termed hybrid Gaussian, which encompasses much fewer free parameters compared to the standard form by exploiting parameter redundancy.
Each Gaussian maintains unique explicit attributes, such as position and opacity, while sharing implicit attributes, including the covariance matrix and view-dependent color, within a latent feature space.

Secondly, we organize hybrid Gaussians in a hierarchical manner, conceptualizing them as a forest structure for scene modeling. Explicit and implicit attributes of hybrid Gaussians are designated as leaf and non-leaf nodes, respectively, and interconnected through efficient pointers. 
In this formulation, through recursive tracing pointers upwards, each leaf node follows a unique path leading to the root of its corresponding tree, and all nodes along this path uniquely characterize a hybrid Gaussian.
In addition, implicit attribute nodes at higher levels are significantly fewer in number 
(the quantity of root nodes is around 2\% of that of the leaves) and are reused by a larger set of hybrid Gaussians, leading to a more compact scene representation while preserving adaptability and expressive capability.

Thirdly, we propose an adaptive growth and pruning strategy for GaussianForest.
This dynamic growth relies on cumulative gradients to discern regions characterized by under-reconstruction or high uncertainty, such as object boundaries or regions with notable view-dependency.  
The expansion of new nodes in these complex regions facilitates swift scene adaptation, even with sparse or imprecise initial points.
Simultaneously, regularly identifying and pruning insignificant leaves and branches, such as trivial Gaussians in simple regions like backgrounds, provides effective control over the total number of nodes. This ensures concise representations without compromising rendering quality, while contributing to the acceleration of both training and rendering.
The primary contributions of this paper are summarized as follows:
\begin{itemize}


\item Introducing GaussianForest, which represents a scene as a forest composed of hybrid Gaussians. By modeling each Gaussian with significantly fewer parameters, remarkable compactness is achieved while adaptability and expressiveness are retained.

\item Developing adaptive growth and pruning strategies specifically tailored for GaussianForest, facilitating rapid scene adaptation while avoiding unnecessary expansion of the number of Gaussians.

\item Extensive experiments showcase GaussianForest's consistent attainment of comparable rendering quality and speed with a compression rate exceeding $10\times$, strongly affirming its efficacy as an efficient technique for scene representation.

\end{itemize}

\section{Related Work}
Neural Radiance Field (NeRF) was introduced by \cite{nerf} as a milestone in scene representation and novel view synthesis. 
It utilizes neural networks to implicitly and continuously model 3D attributes and renders scenes through differentiable ray marching. 
Subsequent work has expanded upon this concept, making significant strides in various areas.
For instance, \cite{adverse1,adverse2,adverse3} focus on enhancing the robustness of NeRF under adverse conditions, such as sparse view configurations and imperfect pose inputs. Meanwhile, \cite{depth1,depth2,depth3,depth4} incorporate depth information as priors to regularize the reconstruction of radiance fields. Furthermore, NeRF's applications have been extended to dynamic scene reconstruction \cite{dynamic1,dynamic2,dynamic3}, scene editing \cite{edit1,edit2,edit3,edit4}, and human head/face modeling \cite{head,head2,face}.
Among all advancements, the evolution in the scene representation and rendering paradigms stand as particularly profound advancements. 
This section provides a brief review of the literature from these two perspectives.

\subsection{Scene Representation}
\subsubsection{Implicit and Explicit Representation}
NeRF \cite{nerf} utilizes MLPs characterized by compact size and continuous mapping to model scenes.
This implicit representation has been embraced by numerous subsequent studies over time. 
However, in addition to the prolonged inference times associated with deep MLPs, updates at arbitrary positions require optimization across the entire network, further exacerbating its inefficiency.
A straightforward acceleration strategy involves a trade-off between space and time: explicitly storing 3D attributes and retrieving them directly. 
Following this principle, Plenoxels \cite{plen} partitions the 3D space and stores the associated attributes within each grid. 
However, high-resolution grids are necessary for detailed rendering, which significantly increases storage requirements. 
TensoRF \cite{tensorf} addresses this issue by applying Tensor Decomposition to 3D grids, substantially reducing the model size, although it remains considerably larger than implicit approaches. 
Meanwhile, 3DGS \cite{3dgs} represents scenes using collections of explicitly represented Gaussians, where millions of Gaussians are required for high-fidelity modeling, resulting in substantial model sizes.

\subsubsection{Hybrid Representation}
Typical hybrid representations involve the explicit storage of implicit features, which are inferred into concrete spatial attributes using neural networks on-the-fly. For example, DVGO \cite{dvgo} stores spatial features within volumetric 3D grids, while Point-NeRF \cite{point} uses discrete 3D points; both subsequently decode these features using MLPs. Generally, hybrid representations combine the flexible nature of implicit representations with the high time efficiency of explicit ones, but they still need to address the storage burden associated with explicit representation. InstantNGP \cite{ngp} demonstrates an effective approach by incorporating multi-resolution 1D hash tables to enable feature sharing among positions with identical hashing values. NSVF \cite{nsvf} uses a sparse voxel octree to define voxel-bounded implicit fields for modeling local properties, enabling faster novel view rendering by skipping empty voxels. Similarly, our GaussianForest is constructed using a hybrid representation of 3D Gaussians and maximizes feature sharing by leveraging spatial redundancy to the fullest extent.




\subsection{Rendering Approach}
\subsubsection{Ray Tracing-based Rendering}
NeRF \cite{nerf} trains and renders scenes via differentiable ray marching, commonly known as volume rendering.
While subsequent research primarily adopted this rendering approach, some work have made their endeavors on enhancing geometry fidelity \cite{volsdf,neus} and improving rendering efficiency \cite{ngp,mip}.
For instance, VolSDF \cite{volsdf} and NeuS \cite{neus} design the transformation between density and signed distance, extending volume rendering to represent SDF (signed distance function) fields for high-quality surface reconstruction.
In pursuit of high efficiency, InstantNGP \cite{ngp} maintains cascade occupancy grids to skip ray marching in empty space, and
Mip-NeRF360 \cite{mip} introduces a proposal network to provide a rapid and approximate scene estimation.
Nevertheless, the computationally intensive nature of dense sampling in ray tracing continues to challenge achieving real-time rendering capabilities.


\begin{figure*}[ht]
    \centering
    \includegraphics[width=1.0\linewidth]{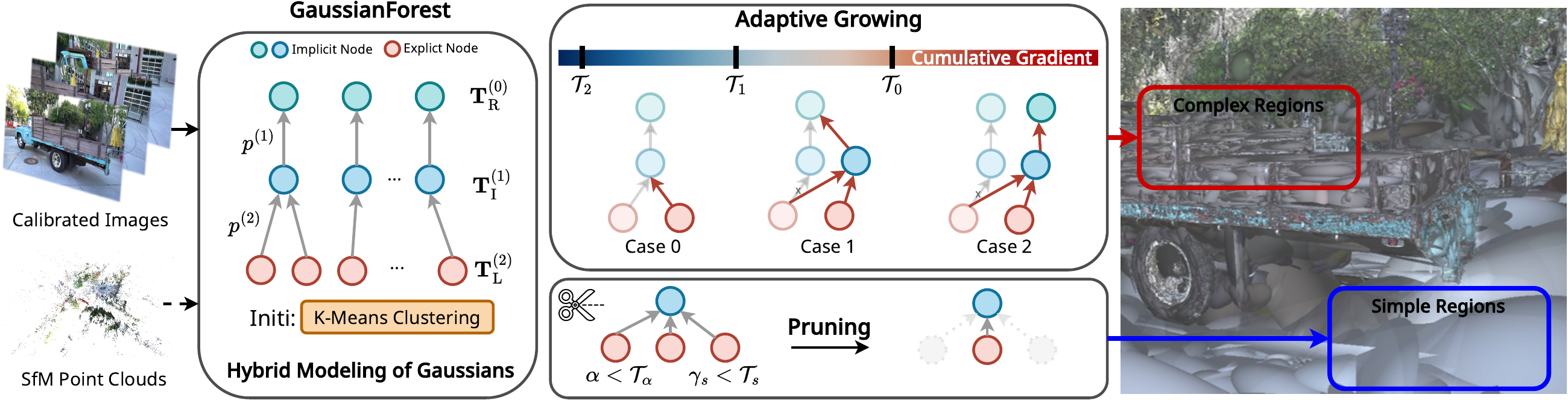}
    \caption{
    \textbf{Illustration of the proposed GaussianForest}. GaussianForest hierarchically represents a scene as a forest composed of hybrid Gaussians, where non-leaf nodes capture their implicit attributes, while leaf nodes characterize explicit ones. Initiated from a compact set of singly linked lists via K-Means, GaussianForest adaptively grows in complex regions based on cumulative gradients to swiftly fit the scene. Leaf nodes with scaling and opacity below certain thresholds are considered trivial and subsequently removed. Such node count control ensures compact representations without compromising rendering quality while contributing to the acceleration of both training and rendering.
}
    \label{fig:overview}
\end{figure*}

\subsubsection{Rasterization-based Rendering}
Recent advancements have propelled differentiable rasterization \cite{rasterizer1,rasterizer2,rasterizer3,rasterizer4,rasterizer5,rasterizer6} into the forefront of computer graphics and computer vision. For instance, \cite{rasterizer1} introduces an approximate gradient solution for differentiable silhouette rasterization, enabling shape reconstruction from silhouette supervision. 
In \cite{rasterizer2}, view rendering is formulated as an aggregation function that fuses the probabilistic contributions of all mesh triangles, facilitating the learning of full mesh attributes from color supervision. 
Diverging from these polygon mesh-based approaches, Pulsar \cite{rasterizer3} introduces a 3D sphere-based differentiable rasterizer that achieves unprecedented speed while avoiding topology problems.
Inspired by Pulsar \cite{rasterizer3}, 3DGS \cite{3dgs} further improves on the concept by employing anisotropic 3D Gaussians instead of isotropic spheres and a rasterizer that respects visibility ordering. 
The fast rendering speed and ease of integration with modern hardware make rasterization a promising avenue for further research. 
Our GaussianForest also follows this highly efficient rasterization approach.

\section{Method}
\subsection{Preliminaries and Task Formulation}
3D Gaussian Splatting (3DGS) \cite{3dgs} aims to model a 3D scene with a set of $N$ anisotropic 3D Gaussians $\{\mathbf{G}_i\}_{i=1}^N$, which are initialized from structure-from-motion (SfM) sparse point clouds. Mathematically, each Gaussian is determined by the point mean position $\bm{\mu}$ and covariance matrix $\mathbf{\Sigma}$ in 3D space:
\begin{equation}
    \mathbf{G}_i(\bm{x}) = e^{-\frac{1}{2}(\bm{x} - \bm{\mu})^{\top}\mathbf{\Sigma}^{-1}(\bm{x} - \bm{\mu})},
\end{equation}
where $\bm{x}$ denotes an arbitrary point position. To ensure the covariance matrix is positive semi-definite throughout the optimization, it is formulated as $\mathbf{\Sigma} = \mathbf{RSS}^{\top}\mathbf{R}^{\top}$ with a rotation matrix $\mathbf{R}$ and a scaling matrix $\mathbf{S}$. 
Specifically, each Gaussian is explicitly parameterized with a group of parameters $\mathbf{\Theta}$:
\begin{equation}\label{eq:param}
    \mathbf{\Theta}:= \{\bm{\mu}, (\mathbf{q}, \mathbf{s}), \alpha, \mathbf{c}\} \in \mathbb{R}^{59},
\end{equation}
where $\mathbf{q}\in\mathbb{R}^4, \mathbf{s}\in\mathbb{R}^3$ are covariance-related quaternion and scaling vectors. $\alpha\in\mathbb{R}$ stands for the opacity for the subsequent blending process. To account for color variations with viewing angles, each Gaussian's color is modeled by 4-order spherical harmonics (SH), represented as $\mathbf{c} \in \mathbb{R}^{3 \times 4^2}$.

After training, the Gaussian parameters are determined, thus allowing the acquisition of the transformed 2D Gaussian on the image plane. 
Subsequently, a tile-based rasterizer is applied to sort $N$ Gaussians for $\alpha$-blending. 
Typically, modeling a real-world scene may require several million Gaussians. 
This substantial quantity, along with the 59 parameters associated with each Gaussian, significantly increases the storage requirements for the trained model and affects rendering speed during the sorting and blending process.

To propose a more streamlined and practical solution, we propose GaussianForest, as illustrated in Fig. \ref{fig:overview}, which models Gaussian parameters within a hybrid tree, where each leaf node traces a distinct path to the root, thereby determining a specific 3D Gaussian. Explicit attributes including position and opacity are stored in the leaf nodes, while implicit attributes like covariance and color are learned in the internal and root layers to maximize sharing across trees and reduce parameter redundancy (Sec. \ref{sec:GaussianForest}). 
Furthermore, to minimize the required number of Gaussians without compromising accuracy, the forests are dynamically grown and pruned (Sec. \ref{sec:growth_pruning}).
\subsection{GaussianForest Modeling}
\label{sec:GaussianForest}
The objective of building the GaussianForest structure is to enable individual Gaussians to retain their unique \textit{explicit} properties for capturing distinct local areas, while simultaneously sharing common \textit{implicit} attributes across the scene to reduce the number of free parameters represented in Eq. \eqref{eq:param}.  
Towards this, we realize this structure through a tree composed of $L$ layers, with $L$ set to 3 without loss of generality. Higher levels of trees will contain fewer nodes with features of higher dimensions to maximize the benefits of sharing.
Herein, we introduce the hybrid modeling of tree nodes.

\subsubsection{Hybrid Tree Nodes} 
The architecture consists of three distinct node types: leaf node $\mathbf{T}_{\text{L}}$ encapsulating explicit attributes; internal node $\mathbf{T}_{\text{I}}$ and root node $\mathbf{T}_{\text{R}}$, both dedicated to implicit attribute modeling. 
Each node type is characterized by a unique set of parameters:
\begin{equation}
\label{eq:tree}
\begin{split}
    &\mathbf{T}_{\text{L}}^{(2)} := \{\bm{\mu}, \gamma_s, \alpha, p^{(2)}\} \in \mathbb{R}^{6},\\
    &\mathbf{T}_{\text{I}}^{(1)} := \{\mathbf{f}_{\text{I}}, p^{(1)}\}\in\mathbb{R}^{\mathcal{D}_{I} + 1}, \\
    &\mathbf{T}^{(0)}_{\text{R}} := \{\mathbf{f}_{\text{R}}\}\in\mathbb{R}^{\mathcal{D}_{R}},
\end{split}
\end{equation}
with the superscript indicating the depth within the tree. 
Among these notations, $\gamma_s$ functions as a scaling coefficient for the implicit scaling $\mathbf{s}$, $p$ represents an integer index pointer, and $\mathbf{f}$ signifies a feature vector, all of which will be further elaborated upon in the following.
Note that such a structure can easily be extended by adding more internal layers.

The rationale behind this hybrid modeling of Gaussian parameters lies in the observation that attributes like position $\bm{\mu}$ and opacity $\alpha$ exhibit sensitivity to deviations, are invariant to viewing angle, and can vary significantly across neighbors. 
Consequently, we classify these as explicit attributes and model them directly in the leaf nodes. In contrast, attributes like covariance-related rotation $\mathbf{q}$, scaling $\mathbf{s}$, and view-dependent colors $\mathbf{c}$ encompass more parameters but typically exhibit local smoothness, leading to redundancy when modeled explicitly. This insight motivates us to model these attributes implicitly, assigning significantly fewer nodes at higher hierarchical levels, i.e., $N_{\text{R}} \ll N_{\text{I}} \ll N$, where $N_{\text{R}}$ and $N_{\text{I}}$ denote the number of root and internal nodes, respectively.
All these number are adaptively adjusted during the forest's growth and pruning processes (Sec. \ref{sec:growth_pruning}).

\subsubsection{Node Traversal} 
Under the formulation of Eq. \eqref{eq:tree}, the pointer $p^{(l)}$ stored at node $\mathbf{T}^{(l)}$ in the $l$-th layer establishes a link from $\mathbf{T}^{(l)}$ to its parent node at the ($l-1$)-th layer. 
By recursively tracing pointers upwards, each leaf node $\mathbf{T}_{\text{L}}^{(2)}$ traverses a unique path $[\mathbf{T}_{\text{L}}^{(2)}, \mathbf{T}_{\text{I}}^{(1)}, \mathbf{T}_{\text{R}}^{(0)}]$ leading to the root $\mathbf{T}_{\text{R}}^{(0)}$ of its corresponding tree, collectively defining the parameters of a distinct 3D Gaussian. 
For the representation of implicit attributes, we concatenate the latent features along this path as $\mathbf{f} = [\mathbf{f}_{\text{I}}, \mathbf{f}_{\text{R}}]\in\mathbb{R}^{\mathcal{D}_{I} + \mathcal{D}_{R}}$.

\subsubsection{Implicit Attributes} To model the implicit attributes, we employ two MLP $\mathcal{F}_{\text{cov}}$ and $\mathcal{F}_{\text{rgb}}$ to decode the obtained latent features $\mathbf{f}$. This decoding results in: (1) the covariance-related scaling $\mathbf{s}$ and rotation vectors $\mathbf{q}$, and (2) the view-dependent color $\mathbf{c}$, as shown below:
\begin{equation}\label{eq:decode1}
\begin{split}
    &\mathbf{s} = \gamma_s \sigma(\mathbf{\hat{s}}),~ \mathbf{\hat{s}}, \mathbf{q} = \mathcal{F}_{\text{cov}}(\mathbf{f}),
\end{split}
\end{equation}
where $\sigma$ indicates a sigmoid activation. The color decoding process additionally incorporates the viewing direction $\vec{\mathbf{d}}$ of the camera as input:
\begin{equation}
    \mathbf{c} = \mathcal{F}_{\text{rgb}}(\mathbf{f}, \vec{\mathbf{d}}).
\end{equation}
This hierarchical representation offers efficiency advantages by reducing over 80\%  of parameters while retaining adaptability and expressive capabilities. Detailed theoretical and empirical analyses regarding storage size are presented in Sec. \ref{sec:complexity}. In conclusion, Gaussian parameterization in Eq. \eqref{eq:param} can be rewritten as:
\begin{equation}\label{eq:hybrid}
    \mathbf{\Theta}_{\text{GF}} := \{\bm{\mu}, \mathcal{F}_{\text{cov}}(\mathbf{f}; \gamma_s), \alpha, \mathcal{F}_{\text{rgb}}(\mathbf{f}, \vec{\mathbf{d}})\}. 
\end{equation}
This formulation minimizes the number of parameters for each Gaussian component. However, this minimization is effective only when the total number of nodes is controlled. In the subsequent section, we elaborate on strategies for optimizing the GaussianForest structure.

\subsubsection{Motivation Illustration}

\begin{figure}[h]
    \centering
    \includegraphics[width=1\linewidth]{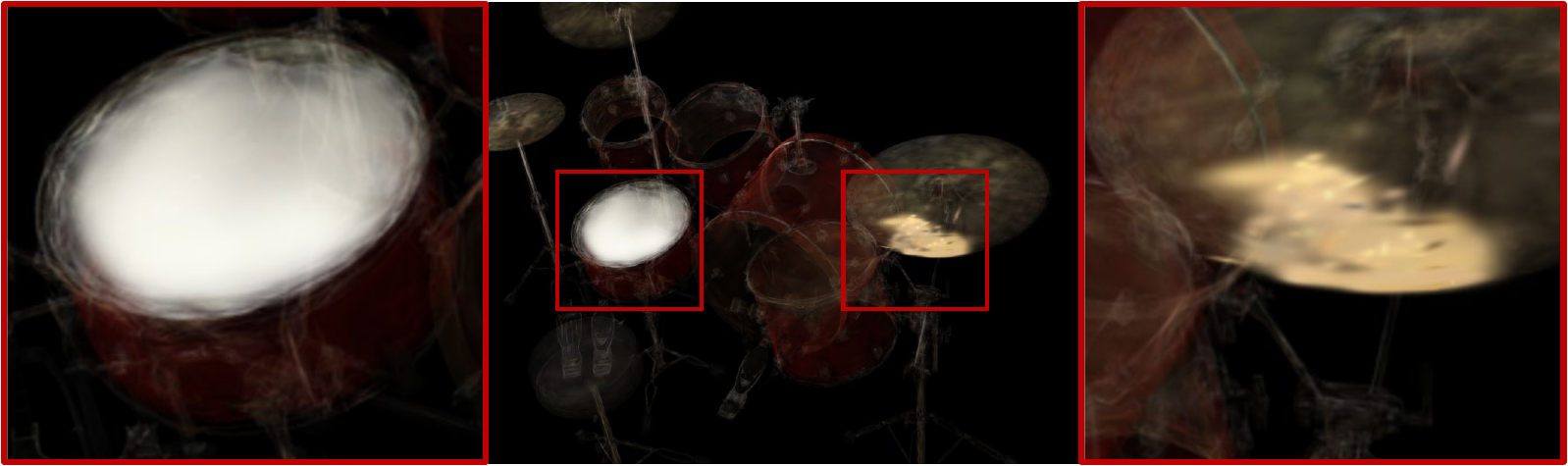}
    \caption{
    Visualization of two Gaussian trees.
    A drum head is approximately modeled by a single Gaussian tree, requiring far fewer parameters compared to the thousands of explicit Gaussians in 3DGS.
    GaussianForest also shows inherent clustering ability, naturally segmenting similar regions without any supervisory information. 
    During optimization, adjacent regions with similar geometric and color features tend to aggregate under the same parent node. 
}
    \label{fig:motivation}
\end{figure}

As previously mentioned, our motivation arises from the high similarity among local Gaussians, which results in substantial parameter redundancy when modeled individually, as shown in 3DGS \cite{3dgs}. Fig. \ref{fig:motivation} exemplifies this insight and intuitively demonstrates the efficiency and potential of GaussianForest. In this figure, we visualize two Gaussian trees by reducing the opacity of the others for clear observation. Specifically, we maintain the opacity of all Gaussians associated with two selected roots while reducing the opacity of the remaining ones to one-tenth.

As illustrated in Fig. \ref{fig:motivation}, one Gaussian tree effectively models a drum head, while the other represents a portion of a cymbal. It is noteworthy that without this hierarchical-hybrid design, modeling a drum head would necessitate thousands of Gaussians, each with 59 free parameters. 
With GaussianForest, the same task is accomplished using only one root and dozens of internal nodes, along with lightweight leaves with only 6 parameters, thereby significantly reducing the number of required parameters.

Another interesting observation from Fig. \ref{fig:motivation} is that Gaussians belonging to the same parent node exhibit similar positional, geometric, and color characteristics. 
In other words, during the optimization of the GaussianForest, adjacent regions with similar geometric and color features tend to aggregate under the same parent node. 
This inherent clustering ability naturally segments similar regions without any supervisory information. 
As illustrated, GaussianForest approximately delineates a whole drum head, from which we observe the potential of leveraging GaussianForest for unsupervised 3D segmentation and scene understanding.

\begin{table*}
  \centering
  \caption{Quantitative comparisons on 
  Mip-NeRF360, Tanks\&Temples, Deep Blending and Synthetic Blender datasets. 
  The best and second-best outcomes are shown in bold deep blue and light blue, respectively. 
  All scores for compared methods are sourced from published papers or released pre-trained models, except for the hyphen (-) indicating no valid data and $^\dagger$ signifying re-evaluation on our machine.}
  \resizebox{\linewidth}{!}{
  \begin{tabular}{l ccccccc ccccccc}
    \toprule
     \multirow{3}{*}{Method}  & \multicolumn{7}{c}{\textbf{Mip-NeRF360}}           & \multicolumn{7}{c}{\textbf{Tanks\&Temples}}  \\
      \cmidrule(lr){2-8} \cmidrule(lr){9-15} &Train& FPS  & Size  & FPS/MB & SSIM$^\uparrow$  & PSNR$^\uparrow$  & LPIPS$^\downarrow$
            &Train& FPS  & MB  & FPS/MB & SSIM$^\uparrow$  & PSNR$^\uparrow$  & LPIPS$^\downarrow$ \\
    \midrule
Plenoxels \cite{plen} & 26 m & 6.79 & 2150 & 0.003  & 0.626 & 23.08 &  0.463
            & 25 m & 13.0 & 2355 & 0.006 & 0.719 & 21.08  &  0.379\\

NGP-Base \cite{ngp}   & 6 m & 11.7 & 13 & {0.900}  & 0.671 & 25.30  & 0.371    
            & 5 m & 17.1 & 13 & {1.315}  & 0.723 & 21.72  & 0.330    \\

NGP-Big \cite{ngp}    & 8 m & 9.43 & 48 & 0.196  & 0.699 & 25.59  & 0.331    
            & 7 m& 14.4 & 48 & 0.300  & 0.745 & 21.92  &  0.305  \\

Mip-360 \cite{mip}    & 48 h & 0.06 & 8.6 & 0.007  & 0.792 & \textcolor{high}{\textbf{27.69}} & 0.237   
            & 48 h & 0.14 & 8.6 & 0.016 & 0.759 & 22.22 & 0.257   \\
3DGS \cite{3dgs}        & 42 m & 134  & 734 & 0.183  & 0.815 & 27.21 & 0.214 
            & 27 m & 154  & 411 & 0.375  & 0.841 & 23.14 & 0.183 \\

    \midrule
3DGS$^\dagger$ \cite{3dgs}             & 28 m & 105 & 827 & 0.127 & \textcolor{high}{\textbf{0.816}} & \textcolor{mid}{\textbf{27.45}} & \textcolor{high}{\textbf{0.201}}
                             & 16 m & 143 & 454 & 0.315 & \textcolor{high}{\textbf{0.848}} & \textcolor{high}{\textbf{23.73}} & \textcolor{high}{\textbf{0.179}} \\
GF-Large    & 28 m & 105  & 85 & \textcolor{mid}{\textbf{1.235}}  & \textcolor{mid}{\textbf{0.803}} & \textcolor{mid}{\textbf{27.45}}   & \textcolor{mid}{\textbf{0.212}}
            & 16 m & 164  & 45 & \textcolor{mid}{\textbf{3.644}}  & \textcolor{mid}{\textbf{0.839}} & \textcolor{mid}{\textbf{23.67}}   &  \textcolor{mid}{\textbf{0.188}}\\
            
GF-Small    & 26 m & 121  & 50 & \textcolor{high}{\textbf{2.426}}  & {0.797} & 27.33 & 0.219 
            & 15 m & 175  & 38 & \textcolor{high}{\textbf{4.605}}  & {0.836} & {23.56} & 0.194\\
   \midrule
    
    \multirow{3}{*}{Method} & \multicolumn{7}{c}{\textbf{Deep Blending}}         & \multicolumn{7}{c}{\textbf{Synthetic Blender}} \\
     \cmidrule(lr){2-8} \cmidrule(lr){9-15} &Train& FPS  & MB  & FPS/MB  & SSIM$^\uparrow$  & PSNR$^\uparrow$ & LPIPS$^\downarrow$  
            &Train& FPS  & MB  & FPS/MB & SSIM$^\uparrow$  & PSNR$^\uparrow$ & LPIPS$^\downarrow$\\
    \midrule
Plenoxels \cite{plen}  & 28 m& 11.2 & 2765 & 0.004   & 0.795 & 23.06 & 0.510
            & 11 m&  -    & 778  &  -      & 0.958 & 31.71 & - \\

NGP-Base \cite{ngp}    & 7 m & 3.26 & 13 & {0.251}  & 0.797 & 23.62 & 0.423  
        & 5 m&  -    & 13 &    -    & 0.963 & 33.18 & - \\

NGP-Big \cite{ngp} & 8 m & 2.79  & 48 & 0.058  & 0.817 & 24.96  & 0.390   
        & -     & -  &   -   & -     & -     & -  &   - \\

Mip-360 \cite{mip} & 48 h & 0.09 & 8.6 & 0.010 & 0.901 & 29.40 & 0.245    
        & 48 h &  -   & 8.6 &  -     & 0.961 & 33.09 &  - \\
3DGS \cite{3dgs}    & 36 m &  137  &676 & 0.203 & 0.903 & 29.41 & 0.243 
        & -     & -  &    -   & - & -     & 33.32 &     -  \\

    \midrule
3DGS$^\dagger$ \cite{3dgs}  & 25 m & 106  & 701 & 0.151   & {0.904} & {29.54} & \textcolor{mid}{\textbf{0.221}}
                & 7 m & 344  & 72  & 4.778  & \textcolor{high}{\textbf{0.969}}  & \textcolor{high}{\textbf{33.80}} & \textcolor{high}{\textbf{0.002}}   \\

GF-Large    & 29 m & 96  & 98 & \textcolor{mid}{\textbf{0.980}}  & \textcolor{high}{\textbf{0.908}} &  \textcolor{high}{\textbf{30.18}} & \textcolor{high}{\textbf{0.215}}  
            & 7 m & 417 & 11 & \textcolor{mid}{\textbf{37.91}}  & \textcolor{high}{\textbf{0.969}} &  \textcolor{mid}{\textbf{33.60}} &  \textcolor{high}{\textbf{0.002}}  \\

GF-Small    & 25 m & 107  & 64  & \textcolor{high}{\textbf{1.672}}  & \textcolor{mid}{\textbf{0.905}}  & \textcolor{mid}{\textbf{30.11}}  & 0.223
            & 6 m & 445  & 8.5 & \textcolor{high}{\textbf{52.71}}  & \textcolor{mid}{\textbf{0.967}}  & {33.52}  &  \textcolor{high}{\textbf{0.002}} \\
   \bottomrule
  \end{tabular}}
  \label{table}
\end{table*}

\subsection{Forest Growing and Pruning}\label{sec:growth_pruning}
To control the number of nodes required for scene modeling, GaussianForest is initialized as a small set of \textit{singly linked lists} and undergoes adaptive growth and pruning to evolve into an efficient and robust forest.
Specifically, branches exhibiting underfitting or high uncertainty are selectively expanded by adding more leaf and/or non-leaf nodes. 
In order to prevent excessive growth of the forest, we implement early stopping and pruning strategies. 
These approaches are pivotal in maintaining a concise yet faithful presentation, accelerating both the training and rendering processes.

\subsubsection{Initialization} 
\label{sec:initialization}
With the given SfM point cloud containing $N_{\text{SfM}}$ points, we correspondingly establish $N_{\text{SfM}}$ leaf nodes with explicit attributes initialized according to 3DGS \cite{3dgs}.
Following this, we initialize the root and internal layer, with each layer comprising $K$ nodes where $K\ll N_{\text{SfM}}$. 
Nodes in two consecutive layers are interconnected in a one-to-one manner. 
Subsequently, we employ the K-means algorithm to group the leaf nodes into $K$ clusters based on proximity.
Each leaf node is then connected to an internal node corresponding to its cluster, thereby
forming $N_{\text{SfM}}$ singly linked lists.
All implicit feature vectors in $\mathbf{T}_{\text{I}}^{(1)}$ and $\mathbf{T}_{\text{R}}^{(0)}$ are randomly initialized.  
For scenes like Synthetic Blending \cite{nerf} with no available SfM point clouds, leaf nodes are initialized using $N_{\text{SfM}}=100$k synthetic points generated by uniform sampling following 3DGS \cite{3dgs}.

\subsubsection{Forest Growth}
\label{sec:growth}
To adapt to varying complexities in different scenes, a hierarchical forest growth strategy is leveraged based on cumulative gradients $\mathbf{CG}$ of leaf nodes during the end-to-end optimization. These gradients serve as indicators of the learning difficulty for each Gaussians. Specifically, we consider three distinct cases governed by a set of gradient thresholds denoted by $\{\mathcal{T}_l\}_{l\in [0, L)}$ arranged in non-increasing order, while nodes beyond these cases remain unchanged. 

\noindent \textcolor{seal}{\textbf{Case 0:} $\mathcal{T}_2 < \mathbf{CG} \leq \mathcal{T}_1$.} For leaf nodes satisfying this case, growth is limited to their own cloning, creating a new link to each of their original parent nodes. This aligns with the split strategy in 3DGS \cite{3dgs}.

\noindent \textcolor{seal}{\textbf{Case 1:} $\mathcal{T}_1 < \mathbf{CG} \leq \mathcal{T}_0$.} For leaf nodes satisfying this case, both leaf and their parent internal nodes are cloned, with the original leaf node and its clone redirected toward the newly formed internal node.

\noindent \textcolor{seal}{\textbf{Case 2:} $\mathbf{CG}\geq \mathcal{T}_0$.} For leaf nodes satisfying this case, a complete cloning of all nodes along the paths to their roots is executed, resulting in the formation of a new linked list for each of them. 

All these cases are illustrated in Figure \ref{fig:overview}. 
The motivation behind this hierarchical design is that a minimal $\mathbf{CG}$ of a Gaussian implies the sufficient representational capacity for its local area. Hence, simpler background regions can be effectively modeled with fewer Gaussians. Conversely, a high $\mathbf{CG}$ indicates the need for more detailed features, especially in complex regions like object boundaries or areas varying from different angles. To address this, both the leaves and their parent nodes are replicated, and the original leaves and their clones are then linked to the newly formed non-leaf nodes, thereby enhancing the model's ability to depict these intricate areas with finer detail through increased feature dimensions.

\subsubsection{Early Stopping} To avoid excessive expansion, we restrict forest growth to early stages, gradually stopping the expansion of higher-level nodes and limiting growth to leaf nodes in the final phase. Subsequently, all growth ends, and we concentrate on pruning. This process is regulated by predetermined stopping points ${t}_l$ for each layer, ensuring efficient and targeted development during training.

\subsubsection{Forest Pruning} 
\label{sec:pruning}
Forest growth plays a crucial role in enhancing the model's ability to represent complex regions. 
Nevertheless, this expansion may lead to an excessive increase in the number of both leaf and non-leaf nodes, consequently giving rise to redundant Gaussians.
In response to this challenge, we develop a pruning strategy focused on eliminating redundant and non-essential Gaussians. 
As illustrated in Fig.~\ref{fig:overview}, this strategy involves evaluating the explicit attributes of each leaf node, i.e., scaling vector $\mathbf{s}$ and opacity $\alpha$. 
If these attributes fall below predefined thresholds, denoted as $\mathcal{T}_s$ for scaling and $\mathcal{T}_\alpha$ for opacity, the corresponding leaf nodes are deemed trivial and are thus removed to free up memory. 
The reasoning behind such pruning stems from the observation that Gaussians with minimal contributions to the $\alpha$-blending process, as suggested by their low scaling and opacity values, exert a negligible impact on the model's overall representational quality.
Additional inspections are conducted after each pruning to identify and eliminate nodes with no children.



\begin{figure*}[ht]
    \centering
    \includegraphics[width=1\linewidth]{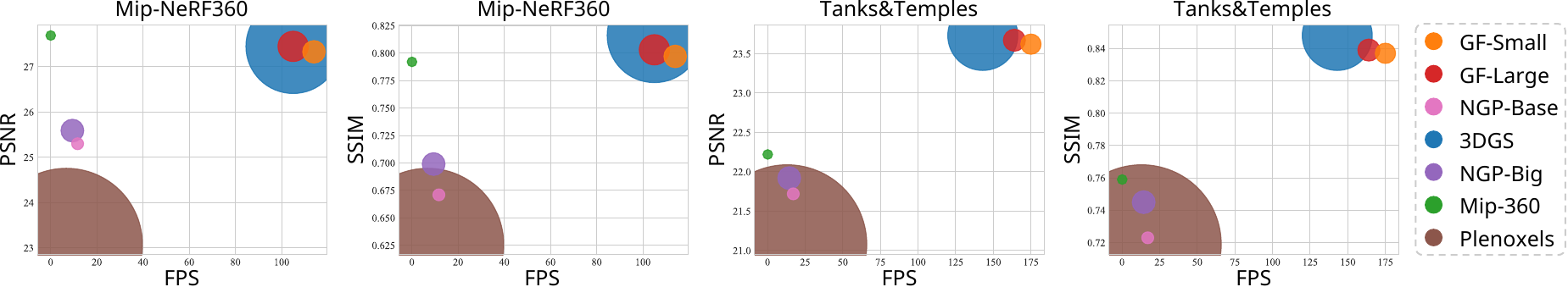}
    \caption{
    Visualization of quantitative comparisons on Mip-NeRF360 and Tanks\&Temples datasets. 
The horizontal and vertical axes represent rendering speed and quality, respectively. Each point's size in the figure indicates the corresponding model size in MB.
This comparison serves to highlight the superiority of our approach.
}
    \label{fig:miptat}
\end{figure*}

\section{Experiments}

\subsection{Implementation Details}

\subsubsection{Framework and Hardware}
Our GaussianForest is implemented based on 3DGS \cite{3dgs} and PyTorch. 
All experiments were conducted on a GeForce RTX 3090 GPU, which shares the same CUDA compute capability (8.6) as the RTX A6000 GPU used in 3DGS \cite{3dgs}. For fair comparison, we re-executed their code on our machine.

\subsubsection{Forest Structure}
We instantiate the GaussianForest as a composition of trees with $L=3$ layers: one root layer, one internal layer, and one leaf layer. Each implicit layer is initialized with $K=10$k nodes, and the feature dimensions for each layer are specified as $\{\mathcal{D}_R, \mathcal{D}_I \} = \{24, 16\}$ and $\{32, 24\}$ for the Small and Large settings, respectively. 

\subsubsection{Forest Growth}
Forest growth occurs every 100 iterations, with growth thresholds set at $\{\mathcal{T}_l\}=\{1 \times 10^{-3}, 2.5 \times 10^{-4}, 2 \times 10^{-4}\}$. 
The number of iterations to stop the growth of each layer is defined as $\{t_l\}$ = \{5k, 10k, 15k\}, and the training process concludes after the 30k iterations.

\subsubsection{Forest Pruning}
Gaussians with $\alpha < \mathcal{T}_\alpha$ or $\gamma_s < \mathcal{T}_s$ are identified and pruned every 100 iterations, where $\{\mathcal{T}_\alpha, \mathcal{T}_s\} = \{1\times 10^{-2}, 5\times 10^{-4}\}$.
In contrast to the early-stop strategy for forest growth, pruning continues until the end of training, with a larger interval defined as 1,000 iterations.

\subsubsection{Features and Decoders}
The two MLPs, $\mathcal{F}_{\text{rgb}}$ and $\mathcal{F}_{\text{cov}}$, are implemented using the fast fully-fused-MLPs from Tiny-CUDA-NN. 
Each MLP consists of 2 hidden layers and is 64 neurons wide.
All features are represented in 16-bit half-float, aligning with the output of the fully-fused-MLPs.

\subsection{Comparative Methods}
3DGS \cite{3dgs} stands out for its SOTA performance in rendering speed and quality, albeit with a substantial model parameter count.
We primarily compare our approach with 3DGS \cite{3dgs}, as we aim to preserve or even enhance the rendering speed and quality while reducing the parameter count.
Additionally, we compared with three representative ray tracing-based radiance field methods, each employing a different type of scene representation: Plenoxels \cite{plen} with explicit scene representation; InstantNGP \cite{ngp} with a hybrid representation; and Mip-NeRF360 \cite{mip} with an implicit one.
These three approaches represent typical examples of different scene representation methods. 
Contrasting with them allows for a comprehensive demonstration of the characteristics of our method.


\subsection{Datasets and Metrics}
Following 3DGS, 
our model has been evaluated across 21 diverse scenarios. 
Of these, 13 scenes are based on real-world captures, including all nine scenes introduced by Mip-NeRF360 \cite{mip}, two scenes from the Tanks\&Temples dataset \cite{tat}, and two from Deep Blending \cite{db}. 
Additionally, all eight synthetic scenes from the Synthetic Blender dataset \cite{nerf} are incorporated. 
These datasets encompass large-scale unbounded outdoor environments, indoor settings, and object-centric scenes.
We employ commonly used PSNR, SSIM \cite{ssim}, and LPIPS \cite{lpips} for evaluating rendering quality. 
Moreover, we provide information on rendering speed, model size, and training time, along with the speed-to-size ratio for a comprehensive and straightforward comparison. 

\begin{figure*}[h]
    \centering
    \includegraphics[width=\linewidth]{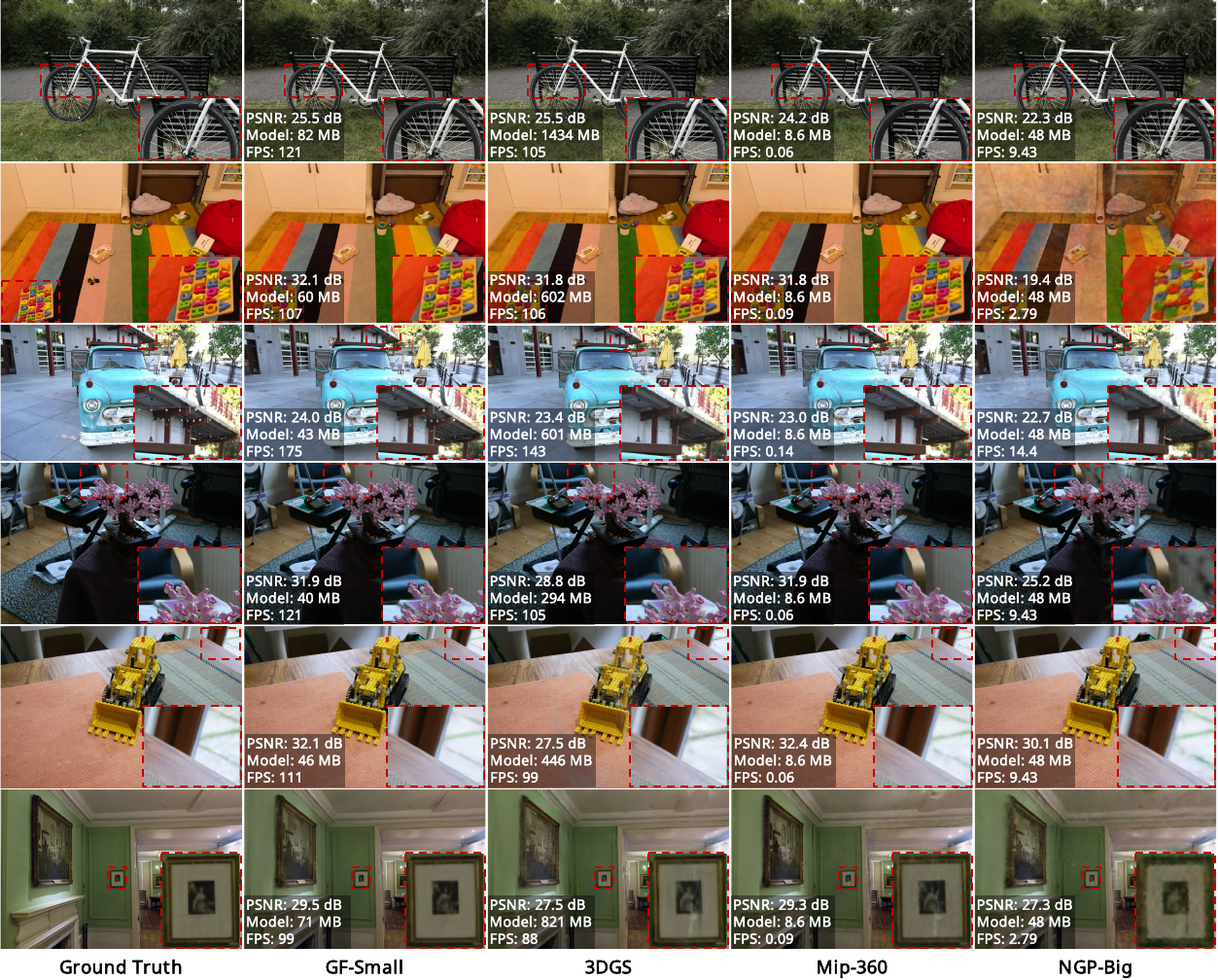}
    \caption{
Qualitative comparisons illustrating rendering quality, with images generated from held-out test views. 
}
    \label{fig:vis}
\end{figure*}

\subsection{Results and Analyses}
\subsubsection{Quantitative Comparisons}


Quantitative results across four benchmarks are presented in Table~\ref{table}, accompanied by additional visualizations for comparisons on Mip-NeRF360 and Tanks\&Temples showcased in Fig.~\ref{fig:miptat}. 
Firstly, our approach excels in adeptly balancing rendering speed and model size. Across all scenarios, GaussianForest achieves the highest speed-to-size ratio, surpassing all comparative methods by a large margin while ensuring high-fidelity rendering quality. 

In addition, compared to the unprecedentedly fast real-time rendering speed achieved by 3DGS, our method not only maintains comparable rendering quality across all test scenarios but also achieves further improvements in rendering speed and training speed. This enhancement is attributed to the substantial reduction in the number of Gaussians facilitated by our adaptive growth and pruning strategies. Most notably, coupled with the efficient scene representation and Gaussian management of GaussianForest, our method achieves a remarkable $7\sim17$ times reduction in model size compared to 3DGS, depending on the dataset and settings.
Beyond the advantages of faster rendering speed and a significantly reduced model size, our approach has remarkably exceeded the rendering quality achieved by 3DGS on Deep Blending. This thoroughly validates the effectiveness of our approach.


\subsubsection{Qualitative Comparisons}
In Fig. \ref{fig:vis}, we present a comprehensive comparison of rendering quality between GaussianForest and 3DGS \cite{3dgs}, as well as representative ray tracing-based methods, including Mip-NeRF360 \cite{mip} and InstantNGP \cite{ngp}.
Across distinct datasets, our findings reveal comparable or even superior quality in the synthesis of novel views. 
This achievement is coupled with the fastest rendering speed, as well as a remarkable compression of parameters exceeding tenfold and notable improvements in both training and rendering speeds compared with the current state-of-the-art 3DGS. 
These outcomes substantiate the efficacy of our proposed method, aligning closely with quantitative results. 

\subsubsection{Complexity Analysis}
\label{sec:complexity}
Assuming 3DGS necessitates $N$ Gaussians for scene modeling, its spatial complexity stands at $O(59N)$. 
In GaussianForest, explicit attributes of each hybrid Gaussian are represented by 6 parameters in its corresponding leaf node, yielding a spatial complexity of $O(6N)$.
Post-adaptive growth, the root and internal nodes account for about $1.5\%\sim2.5\%$ and $25\%\sim50\%$ of leaf nodes, respectively. 
Under the configuration of $\{\mathcal{D}_R, \mathcal{D}_I \} = \{24, 16\}$ and 16-bit half-float feature format, the spatial complexity of non-leaf nodes ranges from $O(2.2N)$ to $O(4.3N)$ (excluding negligible parameters for MLPs and integer pointers in internal nodes). 
Furthermore, our pruning strategy effectively reduces leaf nodes by $1.5\sim 3$ times and non-leaf nodes by about $1.5$ times.
In the end, the overall space complexity ranges from $O(3.5N)$ to $O(7N)$, yielding a compression factor of approximately $8 \sim 17$, aligning seamlessly with the quantitative results. 
Such a reduction in Gaussian count has also accelerated both training and rendering speed, completely offsetting the time complexity introduced by the inclusion of MLPs.

\section{Ablation Study}\label{sec:ablation}
Ablation studies were conducted on Deep Blending scenes \cite{db} to validate the core components of GaussianForest, including hybrid representation, forest management, and adaptive growth and pruning strategies. We also investigated the impact of various hyperparameters, such as growth and pruning thresholds and the feature dimensions of non-leaf nodes. 

\subsection{Baseline Setup}
\subsubsection{\textup{\textbf{+Hybrid}}} 
This baseline employs the hybrid Gaussian representation defined in Eq. \eqref{eq:hybrid}. Diverging from the management of explicit and implicit attributes via a forest structure, this baseline adopts a straightforward feature association approach, connecting each hybrid Gaussian to its corresponding feature based on its position.
In particular, spatial features are retained within a multi-resolution hash table $\mathbf{H}$ as firstly introduced in \cite{ngp}, with the configuration adhering to its default settings.
The corresponding feature $\mathbf{f}$ of the hybrid Gaussian $\mathbf{\Theta}_{\text{GF}}$ is retrieved by indexing $\mathbf{H}$ with its position $\bm{\mu}$, denoted by $\mathbf{f} = \mathbf{H}_{[\bm{\mu}]}$.
We adjusted the table size $T$ to control the model's capabilities with $T=\{18,19,20,21,22,23\}$.

\subsubsection{\textup{\textbf{+Forest}}} 
This baseline integrates the hybrid Gaussian representation defined in Eq. \eqref{eq:hybrid}, akin to \textbf{+Hybrid}. However, instead of the hashing feature association approach, \textbf{+Forest} organizes these hybrid representations within a forest structure as defined in Eq. \eqref{eq:tree}.
In addition, \textbf{+Forest} follows the initialization configuration of GaussianForest described in Sec. \ref{sec:initialization}, but does not undergo the adaptive growth and pruning procedure. 
Specifically, a forest composed of trees with $L=3$ layers is adopted, with the number of nodes in both the root and internal layers initialized to $K=10$k. No nodes are added or removed during the training process.
To investigate the impact of feature dimensions $\mathcal{D}_R$ and $ \mathcal{D}_I$ on the model's capabilities, we conducted three sets of experiments:
\begin{itemize}
\item  \textbf{+Forest$_A$}: $\{\mathcal{D}_R, \mathcal{D}_I \} = $ \{16, 8\}

\item  \textbf{+Forest$_B$}: $\{\mathcal{D}_R, \mathcal{D}_I \} = $ \{24, 16\}

\item  \textbf{+Forest$_C$}: $\{\mathcal{D}_R, \mathcal{D}_I \} = $ \{32, 24\}
\end{itemize}
\subsubsection{\textup{\textbf{+Growth}}} 
This baseline builds upon \textbf{+Forest$_A$} and additionally incorporates adaptive growth of non-leaf nodes defined in Sect. \ref{sec:growth}.
The only distinction between \textbf{+Growth} and our comprehensive GaussianForest model lies in the exclusion of the pruning procedure. 
Moreover, we delineated four configurations for the growth thresholds $\{\mathcal{T}_l\}$ to quantitatively illustrate their impacts. 
Specifically, \(\mathcal{T}_2\) is consistently set to 2, aligning with the Gaussian densification strategy employed in 3DGS \cite{3dgs}, while \(\mathcal{T}_0\) and \(\mathcal{T}_1\) are empirically chosen from the geometric series values of 10, 5 and 2.5 (all in $\times 10^{-4}$ unit):
\begin{itemize}
\item \textbf{+Growth$_A$}: $\{\mathcal{T}_l\} = $ \{10, 5, 2\}

\item \textbf{+Growth$_B$}: $\{\mathcal{T}_l\} = $ \{10, 2.5, 2\}

\item \textbf{+Growth$_C$}: $\{\mathcal{T}_l\} = $ \{5, 5, 2\}

\item \textbf{+Growth$_D$}: $\{\mathcal{T}_l\} = $ \{5, 2.5, 2\}

\end{itemize}

\subsubsection{\textup{\textbf{+Pruning}}} 
This baseline extends \textbf{+Growth$_B$} since it excels in balancing model size and rendering quality as shown in Fig. \ref{fig:ab}. 
Additionally, the forest pruning strategy defined in Sec. \ref{sec:pruning} is further integrated, forming our comprehensive GaussianForest model. 
To investigate the impact of the scale pruning threshold $\mathcal{T}_s$, we define seven settings, 
labeled \textbf{\textit{A}}, \textbf{\textit{B}}, \textbf{\textit{C}}, \textbf{\textit{D}}, \textbf{\textit{E}}, \textbf{\textit{F}}, and \textbf{\textit{G}}, with $\mathcal{T}_s$ = $\{1, 10, 100, 300, 500, 700, 900$\}, respectively (in $10^{-6}$ units).

\subsection{Results and Analyses}

\begin{figure}[ht]
    \centering
    \includegraphics[width=\linewidth]{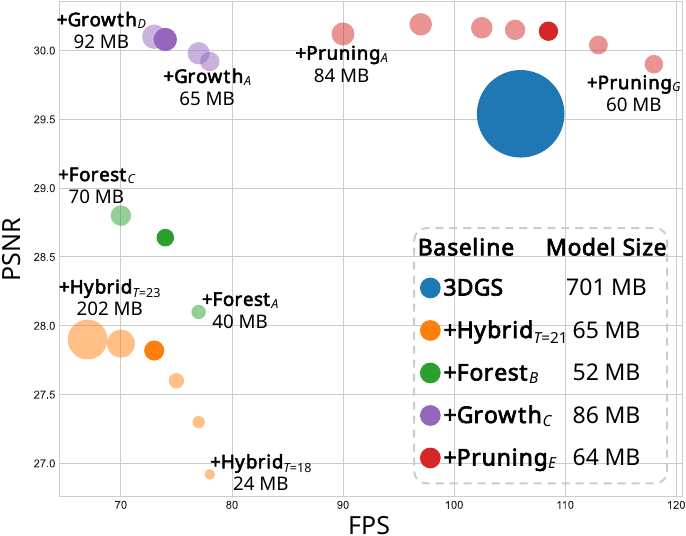}
    \caption{
    Ablation experimental results on Deep Blending scenes. The size of each point in the figure correlates with the respective model size in MB, and 
    each baseline is distinguished by a unique color, with their optimal configurations highlighted in the darkest shade. Furthermore, the legend displays the model sizes corresponding to these optimal settings. To ensure clarity and intuitiveness, only the initial and final points are annotated in the figure for the other parameter settings of each baseline.
}
    \label{fig:ab}
\end{figure}
Fig. \ref{fig:ab} shows that \textbf{+Hybrid}, relying on position-hashing based feature association, achieves a certain level of model compression at a significant cost in rendering quality and speed compared to 3DGS \cite{3dgs}, regardless of the hash table size setting.
By adopting a hierarchical management approach and organizing hybrid Gaussians in a forest structure, \textbf{+Forest} notably enhances rendering quality. 
This highlights the potential of hierarchical-hybrid 3D Gaussian representation for compressed scene modeling.

Nevertheless, the representational capacity of the Gaussian Forest without adaptive growth of feature nodes is severely constrained by the initialization process and feature dimensions. Specifically, leaf nodes initially assigned to the same parent node remain bound together, regardless of how many times they have been cloned. Eventually, the number of leaf nodes may reach millions, while the number of feature nodes remains fixed at the initial $K=10$k. It is evident that the expressive demands cannot be met when hundreds of leaf nodes share a single feature vector in their common parent.

Furthermore, it can be observed that \textbf{+Forest$_B$}, compared to \textbf{+Forest$_A$}, significantly improves rendering quality by increasing the feature dimensions. Although \textbf{+Forest$_C$} further enhances quality, there is a trade-off with decreased rendering speed and memory efficiency. 
Therefore, we choose \textbf{+Forest$_B$} as our primary model configuration (\textit{Small}) and the basis for subsequent ablation studies, reserving \textbf{+Forest$_C$} for the \textit{Large} setting when pursuing higher quality.

It can be observed that by allowing feature nodes to adaptively grow directed by cumulative gradients, \textbf{+Growth} achieves a substantial improvement in rendering quality with a slight increase in parameter count compared to \textbf{+Forest}. 
Notably, its rendering quality even surpasses that of 3DGS \cite{3dgs}, a model nearly ten times larger than ours. This outcome underscores the rationality of our motivation and the effectiveness of our design, i.e., the hierarchical growth of branches in areas characterized by under-reconstruction or high uncertainty. 
This strategy enhances the model's ability to depict intricate areas with finer detail through increased feature dimensions.

Moreover, as the growth threshold gradually decreases from \textbf{+Growth$_A$} to \textbf{+Growth$_D$}, there is an augmentation in the number of feature nodes, coupled with heightened rendering quality, an expanded model size, and a reduction in rendering speed. 
Given the marginal disparity in rendering quality and the trade-off of a smaller parameter count and faster rendering speed between \textbf{+Growth$_C$} and \textbf{+Growth$_D$}, we have chosen \textbf{+Growth$_C$} as our default parameter setting and the foundation for subsequent ablation studies.

From Fig. \ref{fig:ab}, it is evident that introducing the pruning strategy markedly boosts the rendering speed. Given the fact that the total count of Gaussians critically determines the rendering speed by influencing sorting in rasterization and $\alpha$-blending in shading, this advancement is attributed to the effective reduction of the total number of Gaussians compared with \textbf{+Growth} and \cite{3dgs} (approximately half or two-thirds). Moreover, mild pruning not only retains but can slightly improve rendering quality. 
This is because removed Gaussians often contribute insignificantly to the rendering process and scene representation due to their low transparency and scale. Additionally, their removal induces a subtle re-adjustment in the remaining nearby Gaussians, enabling them to compensate for the minor impacts of this elimination through ongoing training.
Ultimately, in light of the balance between rendering quality and time-space efficiency, we have chosen \textbf{+Pruning$_E$} as our definitive GaussianForest model configuration.

\section{Conclusion}
In this paper, we presents a solid solution to the storage issues associated with 3DGS in the context of compressed scene modeling. 
The introduced GaussianForest, with its hierarchical hybrid representation, effectively organizes 3D Gaussians into a forest structure, optimizing parameterization and addressing storage constraints. 
The incorporation of adaptive growth and pruning strategies ensures detailed scene representation in intricate areas while substantially reducing the overall number of Gaussians. 
Through extensive experiments, we demonstrate that GaussianForest maintains rendering speed and quality comparable to 3DGS, while achieving an impressive compression rate exceeding 10 times.


 
%

\bibliographystyle{IEEEtran}
\IEEEtriggeratref{28}
\bibliography{tmm}

\end{document}

%% file: math-extension.tex
\newcommand{\eat}[1]{}

\usepackage{booktabs}
\usepackage{multirow}
\usepackage{siunitx}
\usepackage{makecell, cellspace, caption}
\usepackage{bm}

\usepackage{mathtools}
\usepackage{bbm}
\DeclarePairedDelimiterX{\infdivx}[2]{(}{)}{%
  #1\;\delimsize\|\;#2%
}

\usepackage{stmaryrd}

\makeatletter
\newcommand{\xMapsto}[2][]{\ext@arrow 0599{\Mapstofill@}{#1}{#2}}
\def\Mapstofill@{\arrowfill@{\Mapstochar\Relbar}\Relbar\Rightarrow}
\makeatother

\usepackage{enumitem,kantlipsum}
\usepackage{algorithm}

\usepackage{algpseudocode}

\usepackage{mathrsfs}

\algnewcommand{\Initialize}[1]{%
  \State \textbf{Initialize:}
  \Statex \hspace*{\algorithmicindent}\parbox[t]{.8\linewidth}{\raggedright #1}
}
\algnewcommand{\Inputs}[1]{%
  \State \textbf{Inputs:}
  \Statex \hspace*{\algorithmicindent}\parbox[t]{.8\linewidth}{\raggedright #1}
}
\algnewcommand{\Outputs}[1]{%
  \State \textbf{Outputs:}
  \Statex \hspace*{\algorithmicindent}\parbox[t]{.8\linewidth}{\raggedright #1}
}

\definecolor{seal}{HTML}{B711AC}
\definecolor{maroon}{cmyk}{0,0.87,0.68,0.32}

\usepackage{tikz}

\usepackage{balance}